# Hierarchical Affinity Propagation

**Inmar E. Givoni, Clement Chung, Brendan J. Frey**
Probabilistic and Statistical Inference Group
University of Toronto
10 King's College Road, Toronto, Ontario, Canada, M5S 3G4

## Abstract

Affinity propagation is an exemplar-based clustering algorithm that finds a set of datapoints that best exemplify the data, and associates each datapoint with one exemplar. We extend affinity propagation in a principled way to solve the hierarchical clustering problem, which arises in a variety of domains including biology, sensor networks and decision making in operational research. We derive an inference algorithm that operates by propagating information up and down the hierarchy, and is efficient despite the high-order potentials required for the graphical model formulation.

We demonstrate that our method outperforms greedy techniques that cluster one layer at a time. We show that on an artificial dataset designed to mimic the HIV-strain mutation dynamics, our method outperforms related methods. For real HIV sequences, where the ground truth is not available, we show our method achieves better results, in terms of the underlying objective function, and show the results correspond meaningfully to geographical location and strain subtypes. Finally we report results on using the method for the analysis of mass spectra, showing it performs favorably compared to state-of-the-art methods.

## 1 INTRODUCTION

Exemplar-based clustering (EBC) partitions the data so that each partition is associated with its most prototypical data point ('exemplar'), and the sum of similarities between datapoints and their exemplars and the exemplar points' prior preference to be exemplars is maximized. EBC problems arise naturally in diverse fields and EBC has been used for visual scene analysis (Toyama & Blake, 2002), image segmentation (Xiao et al., 2007), and the analysis of chemical-genetic interaction data and HIV vaccine design (Dueck et al., 2008). EBC is NP-hard, but the affinity propagation (AP) algorithm (Frey & Dueck, 2007) can be used to obtain good quality, albeit approximate, solutions. It is derived as a loopy max-sum (log-domain max-product) algorithm in a factor-graph (Kschischang et al., 2001). Unlike metric-based methods, affinity propagation does not require that the pairwise negative similarities are metrics, and AP has been applied successfully in nonmetric spaces. Unlike non-exemplar methods, such as parametric and nonparametric mixture models (Dempster et al., 1977; Neal, 2000), affinity propagation can use arbitrarily complex similarity functions since it does not need to search or integrate over a parameter space. For example, in (Lindorff-Larsen & Ferkinghoff-Borg, 2009), affinity propagation was used to cluster ensembles of proteins, using the Jensen divergence as the pairwise similarity function.

We extend AP to find hierarchical structure in data. An appealing property of EBC is that a hierarchical representation can be described as an extension of the standard flat representation by finding an exemplar hierarchy. In this representation, for the $l$th layer, only points in the exemplar set of the previous layer $l-1$ may be chosen as exemplars. For example, a hierarchical image segmentation would consist of tiny segments at the lowest layer and successively larger segments at higher layers, corresponding to objects. A greedy approach first described in (Xiao et al., 2007) is to recursively apply AP to the exemplar set found in the previous layer. As we demonstrate in this work, this greedy method suffers by making hard decisions in lower layers that turn out to be suboptimal when higher layers are examined, since it does not optimize a global objective function that accounts for all layers simultaneously.

In this work, we propose a hierarchical exemplar based clustering objective function in terms of a high-order factor-graph, and we derive an efficient approximate loopy max-sum algorithms, which we refer to as hierarchical affinity propagation (HAP). We evaluate the algorithm using different types of synthetic data and show that it achieves better solutions than the greedy alternative that constructs one layer at a time. We also demonstrate the applicability of HAP to two biological tasks: building hierarchies of HIV sequences, and identifying proteins from mass spectra data.

## 2 BACKGROUND

Before deriving HAP in subsequent sections, we first review the standard AP model. For $N$ datapoints, the input is a set of pairwise similarities $\{s_{ij}\}$, where $s_{ij}$ is the similarity of point $j$ to point $i$, and a set of exemplar preferences $\{c_j\}$, where $c_j$ is the preference for choosing point $j$ as an exemplar. The goal is to select a subset of datapoints as exemplars and assign every non-exemplar point to exactly one exemplar, so as to maximize the overall sum of similarities between points and their exemplars and the exemplar preferences. The similarity $s_{ij}$ can be thought of intuitively as the negative Euclidean distance between the datapoints, but it need not be symmetric or metric. Note that in (Frey & Dueck, 2007), $c_j$ was represented by $s_{jj}$; since we explicitly use $c_j$ for the preference, we will assume that $s_{jj} = 0, \forall j$. For the sake of expositional clarity, we assume that all similarities and preferences are negative.

Let $\{h_{ij}\}, i = 1, \ldots, N, j = 1, \ldots, N$, be a set of $N^2$ binary hidden variables where $h_{ij} = 1$ indicates that point $i$ has chosen point $j$ as its exemplar. Furthermore, let $\{e_j\}, j = 1, \ldots, N$, be a set of $N$ binary hidden variables, where $e_j = 1$ indicates point $j$ is chosen as an exemplar. Following (Givoni & Frey, 2009), the objective function for EBC can be expressed using an additive factor graph, as shown in the dotted box in Fig. 1(a), using the following function definitions:

$$C_j(e_j) = c_j e_j, \qquad S_{ij}(h_{ij}) = s_{ij} h_{ij} \qquad (1)$$

$$I_i(h_{i:}) = \begin{cases} 0 & \sum_j h_{ij} = 1, \\ -\infty & \text{otherwise,} \end{cases} \qquad (2)$$

$$E_j(h_{:j}, e_j) = \begin{cases} 0 & e_j = h_{jj} = \max_i h_{ij}, \\ -\infty & \text{otherwise,} \end{cases} \qquad (3)$$

where $h_{:j} = h_{1j}, \ldots, h_{Nj}$ and $h_{i:} = h_{i1}, \ldots, h_{iN}$.

The goal is maximize w.r.t. $\{h_{ij}\}, \{e_j\}$

$$\sum_{i,j} S_{ij}(h_{ij}) + \sum_j \Big(C_j(e_j) + E_j(h_{:j}, e_j)\Big) + \sum_i I_i(h_{i:})$$

The constraint function Eq. (2) forces each point to be assigned to exactly one exemplar (which can be itself), while the function Eq. (3) enforces that $h_{jj} = e_j$ and also enforces 'exemplar consistency', i.e., a point must be an exemplar if other points choose it as an exemplar. This objective can be seen as trading off the exemplar preferences and the pairwise similarities between non-exemplars to exemplars. Thus, the preferences provide the control knob for the total number of clusters to be discovered.

Finding an approximate MAP solution to this NP-hard problem is achieved by running the max-sum algorithm. As was shown in (Frey & Dueck, 2007), the number of binary scalar operations needed in each iteration scales linearly with the number of finite similarities. The two following sets of messages are calculated iteratively until convergence:

$$\alpha_{ij} = \begin{cases} c_j + \sum_{k \neq j} \max(0, \rho_{kj}) & i = j \\ \min[0, c_j + \rho_{jj} + \sum_{k \notin \{i,j\}} \max(0, \rho_{kj})] & i \neq j \end{cases}$$

$$\rho_{ij} = s_{ij} - \max_{k \neq j}(\alpha_{ik} + s_{ik}),$$

where $\alpha_{ij} = 0$ initially. Intuitively, $\rho_{ij}$ corresponds to the extent to which point $i$ wants point $j$ to be its exemplar, while $\alpha_{ij}$ corresponds to how willing point $j$ is to serve as the exemplar for point $i$.

## 3 HIERARCHICAL AP (HAP)

We now show how to generalize the flat AP model to a hierarchical one. We wish to find a set of $L$ consecutive layers of clustering, where the points to be clustered in layer $l$ are constrained to be in the exemplar set of layer $l-1$. This guarantees a hierarchy of exemplars and as we move up in the layers, an exemplar either remains an exemplar or chooses another exemplar to be its exemplar, relinquishing its own role as a cluster representative. Since a greedy layer by layer solution may incur lower preferences for exemplar choices at higher layers due to locally optimal but globally suboptimal decisions made at lower layers, we wish to find a global solution.

Fig. 1(a) describes a graphical model for the hierarchical EBC problem. The main difference compared to the flat representation is manifested in the $I_i^l$ functions:

$$C_j^l(e_j^l) = c_j^l e_j^l, \qquad S_{ij}^l(h_{ij}^l) = s_{ij}^l h_{ij}^l \qquad (4)$$

$$I_i^1(h_{i:}^1) = \begin{cases} 0 & \sum_j h_{ij}^1 = 1, \\ -\infty & \text{otherwise.} \end{cases} \qquad (5)$$

$$I_i^{l>1}(h_{i:}^l, e_i^{l-1}) = \begin{cases} 0 & \sum_j h_{ij}^l = e_i^{l-1}, \\ -\infty & \text{otherwise.} \end{cases} \qquad (6)$$

$$E_j^l(h_{:j}^l, e_j^l) = \begin{cases} 0 & e_j^l = h_{jj}^l = \max_i h_{ij}^l, \\ -\infty & \text{otherwise.} \end{cases} \qquad (7)$$

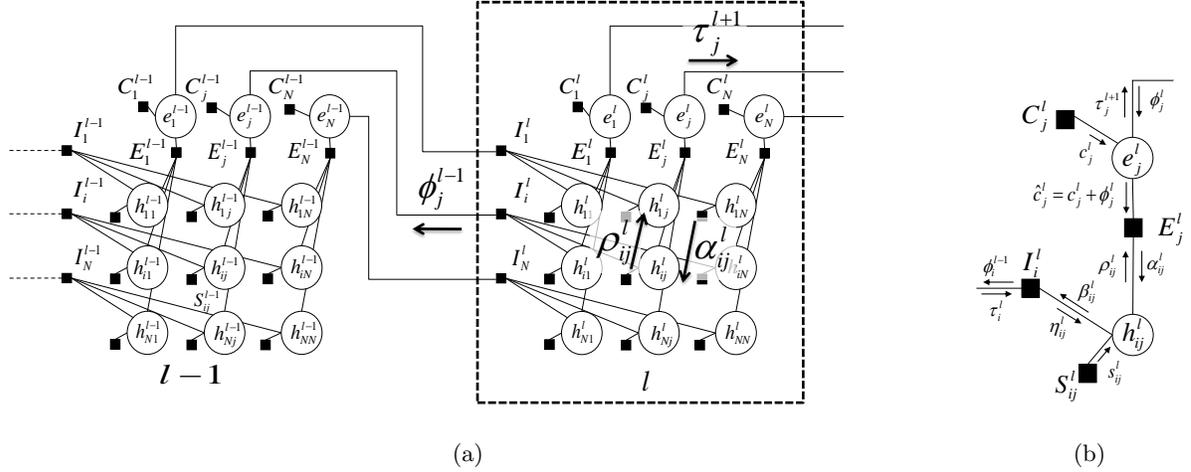

Figure 1: (a) HAP factor-graph, a single layer of the standard AP model is shown in the dotted square. (b) HAP messages.

The modified $I_i^{l>1}$ function results in the following behavior: if point $i$ is *not* chosen as an exemplar at layer $l-1$, (*i.e.* if $e_i^{l-1} = 0$), then point $i$ will *not* be clustered at layer $l$. Alternatively, if point $i$ is chosen as an exemplar at layer $l-1$, it must choose an exemplar at layer $l$. Although we do not explicitly require that only exemplars from previous layers can serve as exemplars at the current layer, this will implicitly be guaranteed through the fact that non-exemplar points in layer $l$ will be constrained to *not* choose an exemplar, together with the exemplar consistency constraint that only allows a point to be an exemplar if it has chosen itself.

The second difference, manifested through the $S_{ij}^l$ and $C_j^l$ functions, is that we allow for layer-specific pairwise similarities ($s_{ij}^l$) and exemplar preferences ($c_j^l$). The rest of the functions remain as before, with the appropriate layer superscripts. As is the case for standard AP, there are no restrictions on the form of the input pairwise similarities and preferences. Thus, the objective function we wish to maximize for hierarchical AP (HAP) can be stated as:

$$\sum_{i,j,l} S_{ij}^l(h_{ij}^l) + \sum_{j,l} C_j^l(e_j^l) + \sum_i I_i^1(h_{i:}^1)$$
$$+ \sum_{i,l>1} I_i^l(h_{i:}^l, e_i^{l-1}) + \sum_{j,l} E_j^l(h_{:j}^l, e_j^l) \qquad (8)$$

For clarity of exposition, we focus here on the formulation and derivation of the hierarchical version for the EBC problem. We note that the EBC problem is highly related to that of the Facility Location (FL) problem, a well studied operational research problem. In FL, the set of exemplars (facilities) is disjoint to the set of datapoints (customers), and the goal is to find a subset of facilities to utilize, and assign each customer to a facility. This relationship was described in (Dueck et al., 2008; Lazic et al., 2009). In addition, it is possible to define an alternative objective function for HAP that makes different restrictions on the exemplars and datapoints – that of clustering all datapoints at every layer, but requiring that the set of exemplars used is nested and consistent across layers.

These related models are formulated via modifications of the layer-specific constraints. This property allows to seamlessly 'mix and match' layers of exemplar-based clustering with facility location in a principled manner, defining novel hierarchical clustering and grouping tasks. The full treatment is given in the supp. material[1].

## 4 HAP ALGORITHM

Similarly to AP, we find solutions to the problem we described above by inferring approximate MAP values for the hidden variables using the max-sum algorithm. Since the function nodes between layers and within layers are high-order, naively, it may seem that they require the evaluation of an exponential number of settings when calculating the maximization needed to compute the max-sum function-to-variable messages. However, much like the case of single layer AP, the set of valid settings for the maximization is restricted and a careful analysis of these valid settings, and shared computations allows us to compute function-to-variable messages efficiently, resulting in runtime of $\mathcal{O}(LN^2)$ per iteration.

The messages to be passed are shown in Fig. 1(b). For an AP layer $l$ using the definition of $I_i^{l>1}$ given in Eq. (6), the messages $\tau_j^{l+1}$ passed up to the next layer from variable node $e_j^l$ to function node $I_j^{l+1}$ and the message $\phi_j^{l-1}$ passed down to the previous layer from

---

[1]http://www.psi.toronto.edu/?q=publications

function node $I_j^l$ to variable node $e_j^{l-1}$ are:

$$\tau_j^{l+1} = c_j^l + \rho_{jj}^l + \sum_{k \neq j} \max(0, \rho_{kj}^l) \quad (9)$$

$$\phi_j^{l-1} = \max_k(\alpha_{jk}^l + s_{jk}^l) \quad (10)$$

Letting $\hat{c}_j^l = c_j^l + \phi_j^l$, the modified $\alpha_{ij}^l$ and $\rho_{ij}^l$ messages for an AP layer are

$$\alpha_{ij}^{l<L} = \begin{cases} \hat{c}_j^l + \sum_{k \neq j} \max(0, \rho_{kj}^l) & i=j \\ \min[0, \hat{c}_j^l + \rho_{jj}^l + \sum_{k \notin \{i,j\}} \max(0, \rho_{kj}^l)] & i \neq j \end{cases} \quad (11)$$

$$\rho_{ij}^{l>1} = s_{ij}^l + \min[\tau_i^l, -\max_{k \neq j}(\alpha_{ik}^l + s_{ik}^l)] \quad (12)$$

We note the $\rho_{ij}^1$ messages passed in the first layer and the $\alpha_{ij}^L$ messages passed in the top-most layer are identical to the standard AP messages for an AP layer.

## 5 EXPERIMENTAL EVALUATION

### 5.1 Synthetic 2D Data

We begin by constructing experiments on synthetic 2D data to determine whether our proposed method, HAP, obtains better results in terms of the objective function we optimize (Eq. (8)), when compared to its greedy bottom-up counterpart used in (Xiao et al., 2007), denoted by 'Greedy'. To examine the applicability of our method under various settings, we look at varying numbers of points and layers. For each such setting we randomly generated increasing exemplar preferences for every layer, and 20 different 2D hierarchical datasets were sampled top-down such that the points of the top-most layer were sampled from a standard normal distribution with a large standard deviation of 3. At consequent layers, sampled points from the previous layer were considered the exemplars. Given those exemplars the layer's points were sampled by randomly picking one of the exemplars, and sampling a point from a normal distribution centered around the exemplar, and with decreasing standard deviation (halved at every layer going down). The total numbers of points sampled at every layer was such that a total overall was of roughly 100, 200, 500, or 750 points across all layers, and such that sampling from top to bottom, every layer contains twice as many points as in the previous layer.

In Fig. 2, we report the median percent improvement over the greedy method for every configuration, aggregating over the 20 different datasets for every configuration, and show scatter plots of the experiments, where each point represents a result for one of the total of 120 experiments carried out for every choice of the total number of points.

We observe that for a larger number of layers (*e.g.* 5-7 layers, compared to 1-4) the improvement of HAP over the greedy method is more pronounced, with improvements of up to 40% for seven layers and 500 datapoints. Further, the bar charts and scatter plots show that as dataset size increases, HAP increasingly outperforms the greedy method. Both of these observations are consistent with the expectation that as the hierarchical clustering problem becomes more difficult (more layers and more datapoints), the gains provided by HAP increase. In a small fraction of cases, HAP does not converge appropriately so that the greedy method achieves better solutions. In practice, a good strategy is to run both methods and pick the solution with the better objective. For these experiments, HAP was on average $\sim 12$ times slower than the greedy method.

For the 2D dataset, we found that HAP tends to oscillate. We therefore use the following procedure to expedite runtime in the synthetic 2D and the real HIV data: we run the algorithm for 500 iterations, fix the variable assignment for the bottom-most layer that has not been fixed yet, and repeat until all layers are fixed. This procedure is used only for the toy-data, and not in the rest of the experiments.

We also investigated the alternative optimization strategy of solving the LP relaxation corresponding to the HAP objective function. We used the provably convergent max-product like MPLP algorithm(Globerson & Jaakkola, 2007) but found the results (objective function value) to be poor in comparison to the other methods. Additionally, the algorithm converges very slowly. See supp. material for further discussion of experimental setup and results.

### 5.2 Analysis of Synthetic HIV Sequences

The process of HIV evolution can be viewed as one of stochastically generating a hierarchy of exemplars, where each exemplar corresponds to an HIV sequence. Given a population of sequences, a useful task is to infer a tree that describes the evolutionary relationships between the sequences. In Sec. 5.3, we use HAP to hierarchically cluster HIV sequences taken from infected individuals. However, since ground truth labelling is not available for that data, we first explore a synthetically generated dataset. Starting with a single root sequence, we simulated the evolution of HIV for three generations to form a four-layer tree of sequences. First, we randomly picked a four-letter root sequence of length 40. To generate children sequences, we first sampled the number of children from a truncated geometric distribution with mean 10. Then, each child was generated by drawing the number of mutations from a geometric distribution with mean 3 and mutating the parent sequence in that many randomly drawn positions. This process was repeated until a four-layer hierarchy was obtained, resulting in a total of 867 sequences. Given the generated sequences, pairwise similarities were obtained using the data-generating mutation process and were set to the logarithm of the truncated geometric distribution evaluated at the observed number of mutations between the sequences.

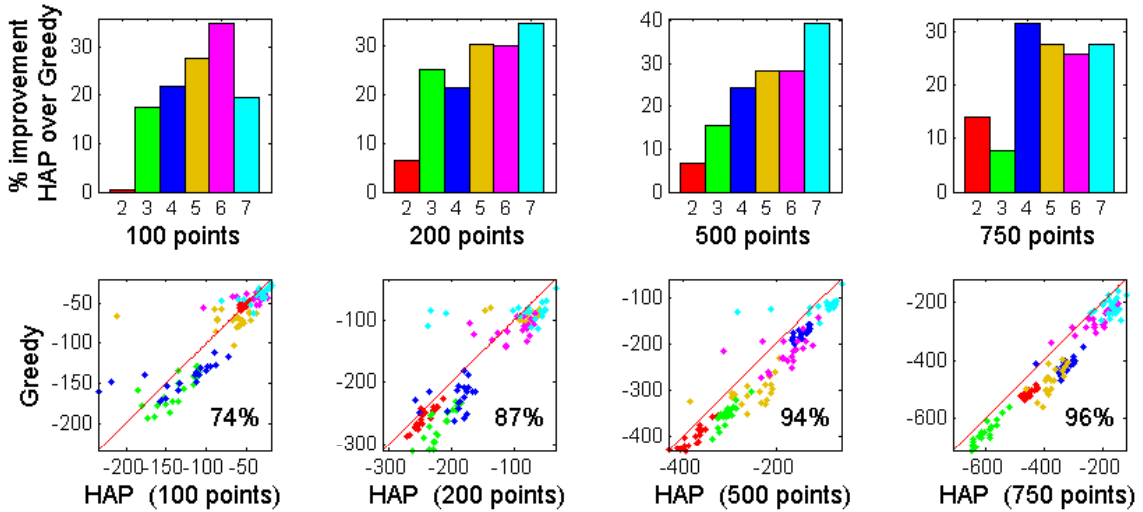

Figure 2: 2D synthetic data: comparison of objective Eq. (8) achieved by HAP and its greedy counterpart (Greedy). Top: Median percent improvement of HAP over Greedy for a given number of layers used. Bottom: Scatter plots of the net similarity achieved by HAP *v.s.* Greedy. Experiments for which HAP obtains better results than Greedy are **below** the line. Total percent of settings where HAP outperforms Greedy is reported in the inset. Color in scatter-plot indicates the number of layers.

We compare the abilities of HAP, its greedy counterpart (Greedy), a hierarchical version of $k$-medians clustering (HKMC), and a hierarchical version of $k$-means clustering (HKMeans) on the task of reconstructing the hierarchy. HKMC works by first identifying exemplars at the lowest layer using $k$-medians clustering (Charikar et al., 2002), and then clustering the resulting exemplars, and so on. For HKMeans, the algorithm finds the lowest layer using $k$-means clustering, and then we post-process the cluster means to find their nearest points.

We applied the different algorithms on a four-layer hierarchy. For HAP and the Greedy, we varied the exemplar preferences to obtain solutions differing in the number of exemplars at different layers; a total of 304 settings were used. For HKMC and HKmeans we ran using a total of 5624 settings, varying the number of clusters per layer, including the ground-truth setting, and for each setting we report the result of the best of 100 random restarts using different initializations for the median or cluster center set.

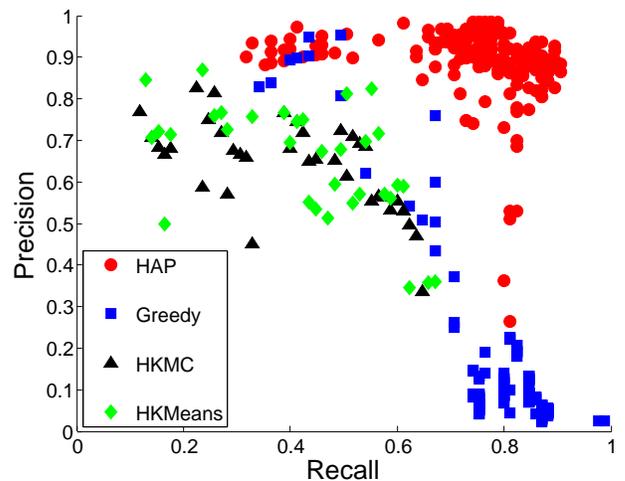

Figure 3: Synthetic HIV data: precision-recall for HAP, Greedy, HKMC and HKMeans applied to the problem of identifying ancestral sequences from a set of 867 synthetic HIV sequences. For HKMC and HKMeans, we only plot the best precision obtained for each unique recall value.

We evaluated the reconstructed trees using two methods. First, we plotted precision *v.s.* recall for various clustering settings (exemplar preferences for HAP and Greedy; random initialization for HKMC and HKMeans), where each sequence-layer combination was labelled as positive if the sequence was generated in that layer in the ground truth, and was labelled as negative otherwise. The precision *v.s.* recall for different methods are plotted in Fig. 3. These results demonstrate that HAP outperforms the other methods at nearly all recall levels. Note that this is also true for the case where HKMC and HKMeans are given the correct parameters for the number of clusters at each layer as input.

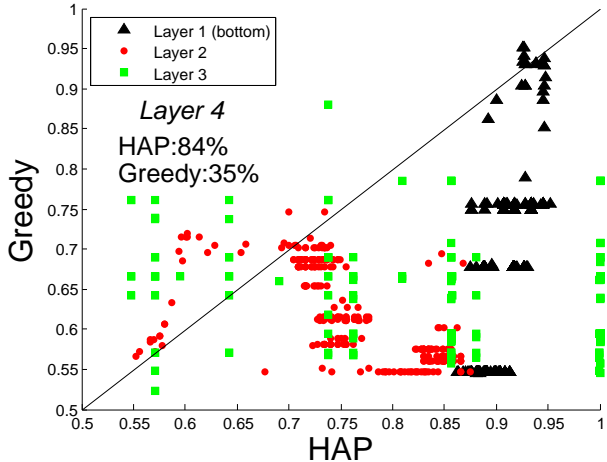

Figure 4: Synthetic HIV data: distribution of Rand index for different experiments using HAP and Greedy. A higher Rand index indicates the solution better resembles the ground truth. Experiments for which HAP obtains better results than Greedy are **below** the line. The percentage of solutions that identified the correct single ancestor sequence at the top layer (layer 4) is also reported.

Additionally, the recall level at which HAP 'breaks' and performance converges to random performance is significantly higher ($\sim 0.82$) than that of the other three methods (Greedy $\sim 0.67$, HKMC & HKMeans $\sim 0.62$ ). Note that applying standard AP layer-by-layer (Greedy) outperforms HKMC, which is expected since, for the single-layer case, AP was shown to outperform $k$-medians clustering. (Frey & Dueck, 2007).

For a more in-depth comparison of HAP and Greedy, we applied an additional evaluation method using modified Rand index ((Rand, 1971; Wagstaff & Cardie, 2000), see supp. material), which measures how frequently pairs of datapoints are correctly classified as siblings or non-siblings by the hierarchical clustering algorithms, given the ground truth clustering. Fig. 4 compares HAP and Greedy for different parameter settings by plotting the Rand index values for each layer in the ground truth labelling.[2] In the high Rand index regime ($>0.8$), HAP almost always performs better than Greedy. It is interesting that by considering all layers, HAP is able to achieve significantly higher Rand indices for the *bottom* layer, as shown by the black triangles. The mean Rand index across all layers and experiments is 0.90 for HAP *v.s.* 0.67 for Greedy. For the first, second and third layers, the mean Rand index is 0.91 for HAP *v.s.* 0.67 for Greedy, 0.76 for HAP *v.s.* 0.62 for Greedy, and 0.86 for HAP *v.s.* 0.64 for Greedy. Since the correct solution has a single ancestral exemplar at the top layer, we calculated the fraction of experiments that correctly identifies the single ancestor. HAP was successful in 84% of the cases, whereas Greedy was successful in only 35% of the cases.

### 5.3 Analysis of Real HIV Sequences

To explore the application of HAP to unlabelled, but real HIV sequences, we obtained 1246 HIV human sequences from the Los-Alamos HIV database[3]. The sequence pairwise similarity was calculated using DNADIST[4], which is a commonly used method for aligning DNA sequences so as to allow for gaps and nonuniform nucleotide substitution. As before, different clusterings were obtained by varying the exemplar preferences.

We compared the performance of HAP to that of Greedy in terms of the objective function in Eq. (8), for a four-layer hierarchy. Exemplar preferences were chosen for each layer at random from the interval $[-0.3, -0.04]$, subject to the constraint that they must decrease as the layer index increases. We obtained a total of 2159 different settings. Results for these experiments are plotted using red discs in Fig. 5 and demonstrate that HAP outperforms the greedy method for all settings. While the ground truth is not known for this data, these results indicate that HAP finds better solutions in terms of the sequence-alignment objective function.

The greatest improvement of HAP over Greedy occurs in the regime of high exemplar preferences across all layers. These are the experimental settings that correspond to the bottom-right area of Fig. 5. A possible explanation for this is that in general, the objective function value at each layer decreases as the number of datapoints decreases. By grouping more datapoints into clusters and finding fewer exemplars at lower layers, fewer datapoints need to be clustered at higher layers. Given high preferences, it might be locally optimal at each layer to have as many datapoints as exemplars as possible. However, globally, it is often better to incur the penalty of clustering more datapoints together at a lower layer and have fewer datapoints in the upper layers to cluster. HAP is able to find these globally better solutions. These are clear cases where a greedy method that optimizes each layer of the hierarchy independently does poorly.

---

[2]This direct comparison on a per-parameter setting basis cannot be replicated for HKMC and HKMeans since they operate on different parameters, namely, the number of clusters to find.

[3]http://www.hiv.lanl.gov/content/index
[4]http://cmgm.stanford.edu/phylip/dnadist.html

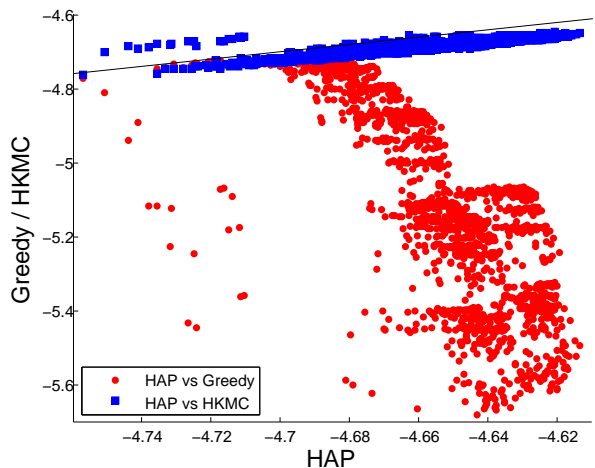
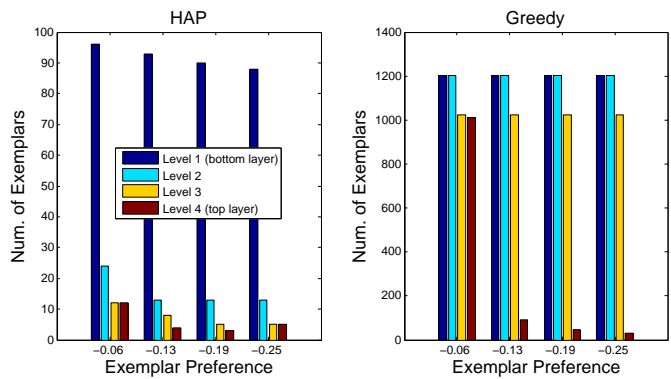

Figure 5: HIV data: comparison of the net similarity achieved by HAP *v.s.* Greedy, and HAP *v.s.* HKMC for a variety of exemplar preference settings. Both axes correspond to net similarity on a log-log scale. In both figures, experiments for which HAP obtains better results than the alternative method are **below** the line.

Figure 6: HIV data: distribution of cluster sizes for various exemplar preferences (model parameter setting) for HAP (left) and Greedy (right). The exemplar preference for layer 4 (top layer) is adjusted as others are held fixed. As exemplar preferences decrease, the overall number of clusters decreases. This has a nonlinear effect on all the layers for HAP but only affects the top layer for Greedy.

To compare HAP with HKMC without the availability of ground truth labelling, we considered the wide range of 2159 clustering solutions found by HAP as described above. For each HAP solution, we determined the number of clusters in each layer and then ran HKMC using that number of clusters, using 100 random restarts. In this way, we were able to compute the net similarity obtained by HAP and HKMC for solutions that had the same number of clusters in corresponding layers. As shown by the blue boxes in Fig. 5, HAP outperforms HKMC in almost all experiments.

To investigate the effect of varying the exemplar preferences in a controlled manner, we fixed the preferences in the bottom three layers and varied the preferences in the top layer. As shown in Fig. 6, when the top-layer exemplar preference decreases, the number of clusters found in the top layer tends to decrease for both HAP and Greedy. However, for HAP, due to the connectivity between layers, a change in the exemplar preference in the top layer affects the number of exemplars found in every layer. For Greedy, each layer is optimized independently, so the top-layer has no effect on the number of exemplars found in lower layers. Thus, even when having many exemplars at the top layer become undesirable, Greedy cannot correct for that by adjusting the lower layers.

Finally, we explored how HAP could be used to partition HIV sequences in interesting ways and compared those results to information about the country of origin of the HIV sequence and its subtype. By tuning the parameter control knobs at different layers, HAP can be used to identify partitions that are highly stable. Such configurations are least sensitive to variations in the way that the pairwise similarities are computed and are thus more likely to be biologically relevant. We identified such a setting of parameters, which corresponded to $(93, 13, 3, 1)$ exemplars (from the bottom-most to the top-most layers).

For each HIV sequence, the Los-Alamos HIV database was used to extract the country of origin (one of 72 countries) and the HIV subtype (one of 113 subtypes), which is a categorization used by HIV researchers that is thought to be related to evolutionary groupings (Robertson et al., 1995). Fig. 7 shows the three third-layer clusters identified by HAP, broken down by their composition of subtype and country of origin. The first two clusters account for 97% of the data, while the third cluster is more specific. It is mostly associated with African strains, and is predominantly of two type 'O' variants, as well as a type 'N' variant. This compares well with focused studies showing that these strains are mostly restricted to west-central African regions (Yamaguchi et al., 2006; Peeters et al., 1997). The first cluster groups many of the 'A' variant subtypes, while the second cluster groups the 'B','C', and 'D' subtypes. The color coding indicates that different clusters correspond to mostly nonoverlapping sets of subtypes. Also, note that South American strains are found only in the second cluster. Analysis of the HIV subtypes found in the second layer of the HAP solution is available in the supp. materials.

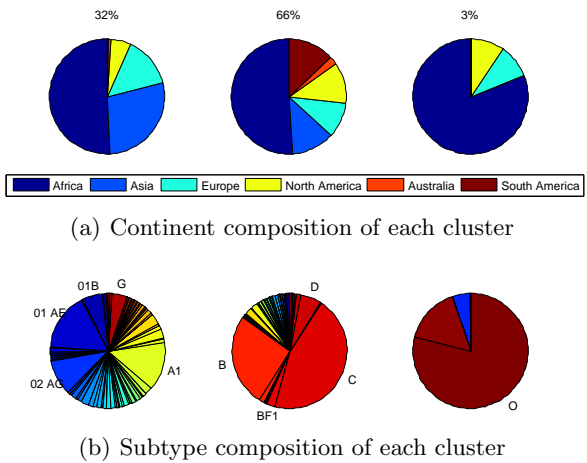

(a) Continent composition of each cluster

(b) Subtype composition of each cluster

Figure 7: HIV Data: third layer. Percentage indicate the number of datapoints associated with each cluster, of all datapoints clustered in this layer. Subtype labels shown only if number of datapoints associated with subtype in cluster is larger than 20.

### 5.4 Protein Prediction from Mass Spectrometry Data

To further demonstrate the applicability of HAP, we used it to identify the protein composition in a biological sample from mass spectrometry (MS) data. In MS, proteins are broken into peptides and their mass spectra are measured and analyzed computationally, to infer the identities of the proteins. However, for complex mixtures, the current state-of-the-art algorithms can only successfully map a small fraction of the mass spectra ($< 20\%$) and so researchers are investigating ways of using prior knowledge about protein groups to improve detection (Keller et al., 2002; Ramakrishnan et al., 2009a,b).

Identifying the proteins can be viewed as a two-layer clustering problem, where mass spectra are assigned to proteins in the bottom layer and proteins are assigned to protein clusters corresponding to functional groups in the top layer (see below). The total score is given by the sum of the mass spectra-protein scores and the protein cluster scores. We include a 'dummy protein' that is used to collect false peptide predictions.

We examined a benchmark dataset consisting of a collection of more than $223,000$ scores (log-probabilities) corresponding to mass spectra-protein pairs for $\sim 6700$ proteins (Eng et al., 1994; Keller et al., 2002; Lu et al., 2006; Ramakrishnan et al., 2009a). Protein cluster scores were extracted from the yeast gene functional network (Lee et al., 2004), which provides similarity scores between pairs of proteins based on various sources of data. These scores were normalized to the interval $(0, 1]$ and then log-transformed. After normal-ization, the only free parameters were the difference between the dummy protein similarities and the similarities for the other proteins, which we varied from $-16$ to $-1$, and the protein cluster preferences, varied from $-5$ to $0$.

We compared our method to three state-of-the-art algorithms: ProteinProphet (Nesvizhskii et al., 2003), MSpresso (Ramakrishnan et al., 2009b) and MSnet (Ramakrishnan et al., 2009a). We used a curated list of proteins from MSnet as a 'gold standard' and compared the ROC curve of our method to the ROC curves of the other methods. Across a wide range of false positive rates, HAP achieves significantly higher true positive rates than ProteinProphet and MSpresso, and performs similarly to the best method from the computational biology literature, MSNet (see also supp. materials). It performs slightly better than a reformulation of the problem as one of facility location, which we solved using message-passing. While these results do not demonstrate that HAP can significantly outperform state-of-the-art computational biology methods, they do indicate the general utility of HAP.

## 6 SUMMARY

We formulated an objective function and described an extended affinity propagation algorithm for hierarchical exemplar-based clustering. This was achieved by generalizing the factor graph and corresponding algorithm that was used to derive the single-layer affinity propagation algorithm. Our method, called hierarchical affinity propagation (HAP), approximately maximizes a natural objective function. By using a message passing method, the algorithm is able to consider all points as potential exemplars at different layers of the hierarchy and send information up and down the hierarchy to identify exemplars at each layer. In this way, HAP is able to outperform methods that use message passing to build the hierarchy layer by layer (Xiao et al., 2007) and techniques that make use of $k$-medians and $k$-means clustering with random restarts. In a small number of cases, HAP fails to properly converge, so a sensible strategy is to apply HAP and a greedy method and pick the one with the best objective function. We also extended HAP to the facility location problem and derived a general framework for constructing hierarchies where each layer can be encoded as either a facility location or an exemplar-based clustering problem. In addition, we found that HAP worked well when it was applied to real-world problems in the field of computational biology, including the analysis of HIV sequences and protein identification. An interesting direction of future research is to apply HAP to problems from other domains such as hierarchical shape matching (Gavrila, 2007) and multi-hop wireless sensor networks (Manjeshwar & Agrawal,

2002), and to investigate decoding schemes with theoretical guarantees as was done for the flat facility location variant of AP (Lazic et al., 2010), an analysis that does not hold for the hierarchial case.

Our main goal in the experimental analysis was to show that given a sensible optimization function, HAP outperforms other variants both in terms of the objective function value and clustering metrics, when ground truth is available, across a wide range of parameter settings. Thus, this makes HAP an algorithm of choice for such problems. The question of how to come up with a good set of parameters was investigated in part in Fig. 6, as well as in our choice of the particular result to analyze in Fig. 7 based on stability of clustering properties such as the number of clusters per layer, and the relative number of clusters per layer being reasonably spread, as we modified the exemplar preference parameters. Further investigations into the problem of explicit parameter setting optimization is an ongoing research direction, which is akin to similar issues that arise both in flat clustering, and hierarchical models such as hierarchical agglomerative clustering, and are known to be hard in the unsupervised clustering setting.